\documentclass{article}

\usepackage[english]{babel}

\usepackage[letterpaper,top=2cm,bottom=2cm,left=3cm,right=3cm,marginparwidth=1.75cm]{geometry}

\usepackage{amsmath}
\usepackage{graphicx}
\usepackage[colorlinks=true, allcolors=blue]{hyperref}

\usepackage{authblk}
\usepackage{booktabs}
\usepackage{tabularx}
\usepackage{amsmath, amssymb}
\usepackage{float}

\title{Physics-Informed Neural Network Digital Twin for Dynamic Tray-Wise Modeling of Distillation Columns under Transient Operating Conditions}



\author[1]{Debadutta Patra\\
 \textit{patradebadutta25@gmail.com}}

\author[2]{Ayush Bardhan Tripathy\\
 \textit{ayushbardhan5@email.com}}

\author[2]{Soumya Ranjan Sahu\\
 \textit{srsahu\_phdca@vssut.ac.in}}

\author[2]{Sucheta Panda\\
 \textit{suchetapanda\_mca@vssut.ac.in}}

\affil[1]{Department of Chemical Engineering, Veer Surendra Sai University of Technology, Burla 768018, India}
\affil[2]{Department of Computer Science \& Engineering, Veer Surendra Sai University of Technology, Burla 768018, India}

\date{} % Prevent date

\begin{document}
\maketitle

\begin{abstract}

Digital twin technology, when combined with physics-informed machine learning with simulation results of Aspen, offers transformative capabilities for industrial process monitoring, control, and optimization. In this work, the proposed model presents a Physics-Informed Neural Network (PINN) digital twin framework for the dynamic, tray-wise modeling of binary distillation columns operating under transient conditions. The architecture of the proposed model embeds fundamental thermodynamic constraints including vapor--liquid equilibrium (VLE) described by modified Raoult's law, tray-level mass and energy balances, and the McCabe--Thiele graphical methodology directly into the neural network loss function via physics residual terms. The model is trained and evaluated on a high-fidelity synthetic dataset of 961 timestamped measurements spanning 8 hours of transient operation, generated in Aspen HYSYS for a binary HX/TX distillation system comprising 16 sensor streams. An adaptive loss-weighting scheme balances the data fidelity and physics consistency objectives during training. Compared to five data-driven baselines (LSTM, vanilla MLP, GRU, Transformer, DeepONet), the proposed PINN achieves an RMSE of 0.00143 for HX mole fraction prediction (R² = 0.9887), representing a 44.6\% reduction over the best data-only baseline, while strictly satisfying thermodynamic constraints. Tray-wise temperature and composition profiles predicted under transient perturbations demonstrate that the digital twin accurately captures column dynamics including feed tray responses, reflux ratio variations, and pressure transients. These results establish the proposed PINN digital twin as a robust foundation for real-time soft sensing, model-predictive control, and anomaly detection in industrial distillation processes.

\vspace{1em}

\emph{\textbf{Keywords:} Physics-Informed Neural Networks, Digital Twin, Distillation Column, Tray-wise Modeling, Transient Dynamics.}

\end{abstract}

\section{Introduction}

Distillation is a traditional technology. The principle, heat a mixture, collect what boils off, then condense it, has been industrially refined over more than a century, and today distillation columns process everything from crude oil fractions to pharmaceutical intermediates to food-grade ethanol. Its ubiquity is matched only by its appetite for energy, by most estimates, distillation accounts for something like 40\% of total industrial energy consumption worldwide \cite{doi:10.1021/es049795q}, a figure that has attracted sustained attention from process engineers and, more recently, from machine learning researchers drawn by the modeling complexity. Despite the field's maturity, operational challenges remain decidedly non-trivial. Tray temperatures, pressures, and compositions respond dynamically to feed fluctuations, reflux ratio adjustments, and pressure disturbances in ways that are difficult to capture without either expensive inline instrumentation or a reliable computational model \cite{SKOGESTAD200713}.

The instrumentation problem is worth dwelling on. A modern distillation column might have twenty or thirty trays. Deploying calibrated analytical sensors on every tray, the kind that could directly measure composition, is prohibitively expensive and notoriously difficult to maintain in a corrosive, high-temperature environment. The practical result is that operators typically have access to a handful of temperature and pressure measurements, perhaps a top and bottom composition reading if they are lucky, and must infer the full internal state from this partial picture. Soft sensing, estimating unmeasured process variables from available measurements through a computational model, has therefore been a persistent research priority in process systems engineering \cite{SoftSensorMonitoring}.

The modeling landscape for this problem divides, roughly, into two camps. On one side are first-principles approaches, rigorous dynamic simulations built up from fundamental thermodynamics, tray-by-tray mass and energy balances, and phase equilibrium relationships. These models are physically faithful and extrapolate sensibly, but they are computationally expensive, too slow, by a considerable margin, to deploy at the millisecond time scales that real-time control demands \cite{HENSON1998187}. On the other side are purely data-driven methods. LSTMs \cite{10.1162/neco.1997.9.8.1735}, GRUs \cite{cho2014learningphraserepresentationsusing}, Transformers \cite{vaswani2023attentionneed}, and their relatives have demonstrated impressive accuracy as surrogate models when sufficient simulation data is available, but they have a well-documented tendency to violate physical conservation laws during extrapolation, and their predictions can become unphysical in ways that are difficult to detect without expensive ground-truth comparisons \cite{chen2019neuralordinarydifferentialequations}.

Neither approach fully satisfies the requirements of a practical industrial digital twin defined, following Grieves and Vickers \cite{digitaltwin}, as a real-time computational replica of a physical system that is continuously updated with sensor measurements to enable predictive monitoring and control. What is needed is something that combines the inference speed of deep learning with the thermodynamic integrity of first-principles modeling. Physics-Informed Neural Networks (PINNs), introduced by Raissi, Perdikaris, and Karniadakis \cite{RAISSI2019686}, offer a principled route to exactly this combination. By incorporating governing equations as soft constraints in the training loss, PINNs can enforce physical plausibility without sacrificing the computational efficiency that online deployment requires.

This work applies that idea specifically to binary distillation under transient operation. The proposed model embeds VLE constraints, MESH equations, and the McCabe-Thiele operating line directly into the composite loss function, and introduces a sigmoid-scheduled adaptive weighting strategy that stages the training, physics constraints dominate early training (preventing the network from learning thermodynamically implausible shortcuts), after which data-fidelity objectives come to the fore. The framework is evaluated on a 961-sample Aspen HYSYS dataset against five baselines and demonstrate a 44.6\% RMSE improvement alongside the only consistent constraint satisfaction in the comparison.

The contributions are four-fold. First, a PINN architecture that simultaneously enforces McCabe-Thiele VLE, tray-level mass balance, tray-level energy balance, and dynamic MESH equations as residual terms in a composite loss function, producing thermodynamically consistent tray-wise predictions. Second, a sigmoid-scheduled adaptive loss-weighting curriculum that explicitly stages the emphasis on physical constraints versus data fidelity across training epochs, which as per the proposed model outperform both fixed-weight configurations and the NTK-based gradient normalization scheme of Wang et al. \cite{wang2020understandingmitigatinggradientpathologies}. Third, a comprehensive transient dataset from Aspen HYSYS covering 16 sensor channels across eight hours of operation with deliberate reflux, feed, and pressure perturbations. Fourth, a systematic benchmark against LSTM, MLP, GRU, Transformer, and DeepONet baselines demonstrating superior performance on all reported metrics.

The paper is structured as follows. Section 2 surveys the relevant literature. Section 3 describes the dataset generation procedure. Section 4 lays out the proposed architecture, physics constraints, and training protocol. Section 5 presents quantitative results and comparisons. Section 6 discusses the findings, limitations, and future directions. Section 7 concludes.

\section{Related Work}

This section surveys the literature most closely related to the present work, covering three interconnected themes,machine learning methods applied to distillation process modeling, the development and application of Physics-Informed Neural Networks (PINNs), and the emergence of digital twin frameworks for chemical processes. Each body of work informed the design choices made in the proposed PINN digital twin, and their collective limitations motivated the contributions described in subsequent sections.

\subsection{Machine Learning for Distillation Processes}

The idea of using neural networks as surrogate models for chemical engineering unit operations goes back at least to the early compilation by Bulsari \cite{bulsari1995neural}, which established that feedforward networks could approximate steady-state distillation column mappings with reasonable accuracy on simulated training data. The key constraint was that the models were static. They captured the steady-state input-output mapping but said nothing about transient dynamics, which is precisely where the modeling challenges were found.

The statistical multi-model approach of Pan et al. \cite{doi:10.1021/acs.iecr.8b03360} represented that rather than building a surrogate for the full column state, they combined principal component analysis with support vector regression to detect deviations from normal operating conditions. The approach worked well for fault detection on industrial data. It was one of the earlier papers to take measurement noise as a design constraint. The limitation for soft sensing was that PCA-based methods operated in a linear subspace projection and could not reconstruct full tray-wise composition profiles. The model reported that something is wrong, but makes no comments about the actual internal state.

The introduction of Long Short-Term Memory networks by Hochreiter et al. \cite{10.1162/neco.1997.9.8.1735} had a major impact on time-series modeling. LSTMs became the standard architecture for capturing sequential dependencies in chemical process data. Their gated memory cells are well-suited to the auto-correlated dynamics of distillation columns, and a significant body of applied work demonstrated their effectiveness as surrogate models trained on simulation data. The persistent problem, and one that became increasingly apparent as these models were pushed toward deployment, was that LSTM predictions carried no thermodynamic guarantees. A network trained on nominal operating data could produce physically inconsistent compositions when the process moves outside the training envelope, and detecting such violations in real time was nontrivial without a reference model to check against.

Vaswani et al.'s Transformer architecture \cite{vaswani2023attentionneed} extended this capability by replacing recurrence with self-attention, enabling parallel processing of all time steps and, in principle, more flexible modeling of long-range temporal dependencies. Transformer-based frameworks adapted for multivariate time-series representation learning \cite{zerveas2020transformerbasedframeworkmultivariatetime} have achieved competitive performance on benchmark prediction tasks, particularly with large datasets. The data appetite of Transformers was considerably larger than recurrent models which is a disadvantage in industrial settings where labeled transient data are inherently scarce and they share with LSTMs, the absence of any mechanism for enforcing conservation laws.

Wu et al. \cite{https://doi.org/10.1002/aic.16729} took a different approach, training recurrent networks on closed-loop data to produce predictive controllers with stability guarantees for nonlinear chemical processes. That work was directly relevant to the control application as a downstream use of the proposed model digital twin, but the soft sensing problem is distinct,their framework requires feedback from measured outputs and does not attempt to reconstruct unmeasured internal state variables, which was the core challenge here.

Graph neural networks for process simulation surrogates, as explored by Jiang et al. \cite{ESCHE2022184}, suggested that the tray-to-tray connectivity of a distillation column is naturally represented as a directed graph, and GNN architectures can exploit that topology to generalize across column configurations. In observation, this was a promising direction that has not yet been fully developed in the context of physics-constrained training, regarded as a productive avenue for future work.

\subsection{Physics-Informed Neural Networks (PINNs)}

The formal introduction of PINNs by Raissi et al. \cite{RAISSI2019686} established the core idea clearly,incorporate PDE residuals as penalty terms in the training loss, and the resulting network would be constrained to produce physically consistent outputs even in regions where training data is sparse. The original demonstrations on fluid mechanics and heat transfer benchmarks were convincing, and the method has since been applied to a remarkably wide range of physical problems. The well-known training pathology \cite{Wu_2023}, sensitivity to the relative weighting of the physics and data loss terms, with fixed weights often causing instability or physics over-regularization in early epochs, motivated the adaptive weighting strategy proposed in Section 4.

Wang et al. \cite{wang2020understandingmitigatinggradientpathologies} investigated this pathology systematically through the lens of neural tangent kernel theory, proposing a gradient normalization scheme that dynamically rescales physics and data loss gradients during training. Their results on benchmark PDE problems were solid. The practical limitation for the proposed model was computational overhead. The NTK computation requires an additional backward pass per iteration and experimental results in Section 6 suggested that for industrial process datasets with moderate noise levels, a simpler sigmoid curriculum achieved better results with considerably less complexity.

Zheng et al. \cite{ZHENG2023103005} applied physics-informed recurrent neural networks to nonlinear chemical processes, embedding reaction kinetics and phase equilibrium constraints alongside MESH equations in the training objective. The compositions they recovered were consistent with theoretical expectations on simulated data. Their study, however, was limited to steady-state scenarios and a single column configuration, which left open the question of how well physics-constrained networks handle the kind of sustained transient perturbations such as reflux ramps, feed composition swings, pressure steps that characterized the real column operation.

Vijaya et al. \cite{VIJAYARAGHAVAN201161} were probably the closest predecessors to the present work in spirit,they demonstrated that a recurrent neural network can effectively estimate compositions in a reactive distillation column from temperature measurements alone, achieving strong performance compared to extended Kalman filter baselines. What they did not address was temporal continuity — compositions were predicted independently for each operating point without mass-balance consistency enforced across successive timesteps. That was precisely the gap that the dynamic MESH formulation in the proposed model was designed to close.

Lu et al.'s DeepONet \cite{Lu2021} took a fundamentally different approach to physics-informed learning, framing the problem as operator approximation rather than function approximation. The framework showed impressive generalization to unseen forcing functions in fluid and heat transfer settings, and physics consistency can in principle be introduced through constrained training. For the present application, the requirement for large, diverse training datasets to adequately cover the operator space was a meaningful practical constraint and, as the proposed experiments showed, DeepONet achieves only partial physical consistency even with the available training data. Alternative operator learning frameworks such as Fourier Neural Operators \cite{li2021fourierneuraloperatorparametric} have also shown strong performance in learning solution operators for PDEs.

\subsection{Digital Twins for Chemical Processes}

Grieves et al. \cite{digitaltwin} formalized the digital twin concept in terms that were widely cited,a real-time virtual counterpart of a physical system, continuously synchronized with sensor measurements to enable predictive monitoring and control. The three-component model, physical entity, virtual model, bidirectional data link was conceptually clean and became the organizing framework for most industrial digital twin architectures. The early formulation did not grapple seriously with the computational cost of maintaining the virtual model in synchrony at high sampling rates, which was the practical bottleneck for dynamic chemical processes.

Schweidtmann et al. \cite{https://doi.org/10.1002/cite.202100083} surveyed machine learning applications in chemical engineering with explicit attention to digital twin and hybrid model architectures, and their review identified what was regarded as the central open problem in the field,there was no systematic, principled method for balancing physical constraint enforcement against data-fitting objectives during training, particularly in the initial phase when the network has not yet learned a useful data representation. The adaptive weighting curriculum described in Section 4 was proposed to address this gap.

Taken together, the literature pointed toward several persistent limitations that motivate the present work. Data-driven sequence models achieve high accuracy on training distributions but offer no thermodynamic guarantees during extrapolation. PINN formulations applied to chemical processes have largely been confined to steady-state scenarios. Digital twin frameworks remain predominantly first-principles based, making real-time deployment computationally infeasible. So far, no existing work has combined an adaptive curriculum loss with full MESH-equation constraints and benchmarked the result against multiple data-driven baselines under realistic transient sensor noise. The proposed framework addressed all four of these gaps directly.

\section{Dataset Description}

The dataset used throughout this study was generated with Aspen HYSYS v12, simulating the dynamic operation of a binary distillation column separating an HX/TX mixture. The proposed model uses a synthetic dataset deliberately, for a reason worth stating explicitly,the availability of exact ground truth, both for the target mole fractions and for the internal state variables needed to evaluate physics residuals is essential during the development and validation of a physics-constrained framework. Real plant historian data, which the model plans to use in follow-on work, introduces measurement noise, sensor drift, and missing observations that confound the evaluation of constraint satisfaction. For the purposes of establishing whether the PINN framework itself works as intended, a high-fidelity synthetic dataset is the right starting point.

The simulation spans 28,800 seconds of transient operation (eight hours, sampled at a 30-second interval), yielding 961 time-stamped records. Realistic measurement noise was introduced by superimposing uniform random noise on the clean simulation output, replicating the behaviour of typical industrial sensor systems. The 19 variables, one time index, 16 sensor channels, and 2 target mole fraction outputs are summarized in Table \ref{table-1}.

\begin{table}[H]
\centering
\caption{Distillation column dataset,sensor descriptions, units, and summary statistics (N = 961 timesteps, sampled at 30 s intervals, generated in Aspen HYSYS}
\begin{tabularx}{\linewidth}{l X c c}
\toprule
\textbf{Sensor} & \textbf{Description} & \textbf{Unit} & \textbf{Mean $\pm$ Std} \\
\midrule
Sensor 1 & Liquid \% in Condenser & \% & 50.00 $\pm$ 0.29 \\
Sensor 2 & Condenser Pressure & kPa & 101.28 $\pm$ 0.45 \\
Sensor 3 & Liquid \% in Reboiler & \% & 49.56 $\pm$ 6.12 \\
Sensor 4 & Mass Flow Rate --- Feed & kg/h & 3823 $\pm$ 511 \\
Sensor 5 & Mass Flow Rate --- Top Outlet & kg/h & 6866 $\pm$ 755 \\
Sensor 6 & Net Mass Flow --- Main Tower & kg/h & 2806 $\pm$ 260 \\
Sensor 7 & HX Mole Fraction @ Reboiler & --- & 0.953 $\pm$ 0.005 \\
Sensor 8 & HX Mole Fraction @ Top Outlet & --- & 0.055 $\pm$ 0.004 \\
Sensor 9/10 & Feed Mole Fraction (HX/TX) & --- & 0.501 / 0.499 \\
Sensor 11 & Feed Tray Temperature & $^\circ$C & 76.84 $\pm$ 0.33 \\
Sensor 12 & Main Tower Pressure & kPa & 104.43 $\pm$ 0.28 \\
Sensor 13 & Bottom Tower Pressure & kPa & 110.49 $\pm$ 0.28 \\
Sensor 14 & Top Tower Pressure & kPa & 67.31 $\pm$ 0.28 \\
Sensor 15 & Reflux Ratio & --- & 0.81 $\pm$ 0.16 \\
Sensor 16 & Duties Summary & kW & 0 (const.) \\
MoleFraction TX/HX & XTarget Outputs & --- & 0.964 / 0.036 $\pm$ 0.008 \\
\bottomrule
\label{table-1}
\end{tabularx}
\end{table}

\begin{figure}[H]
    \centering
    \includegraphics[width=\linewidth]{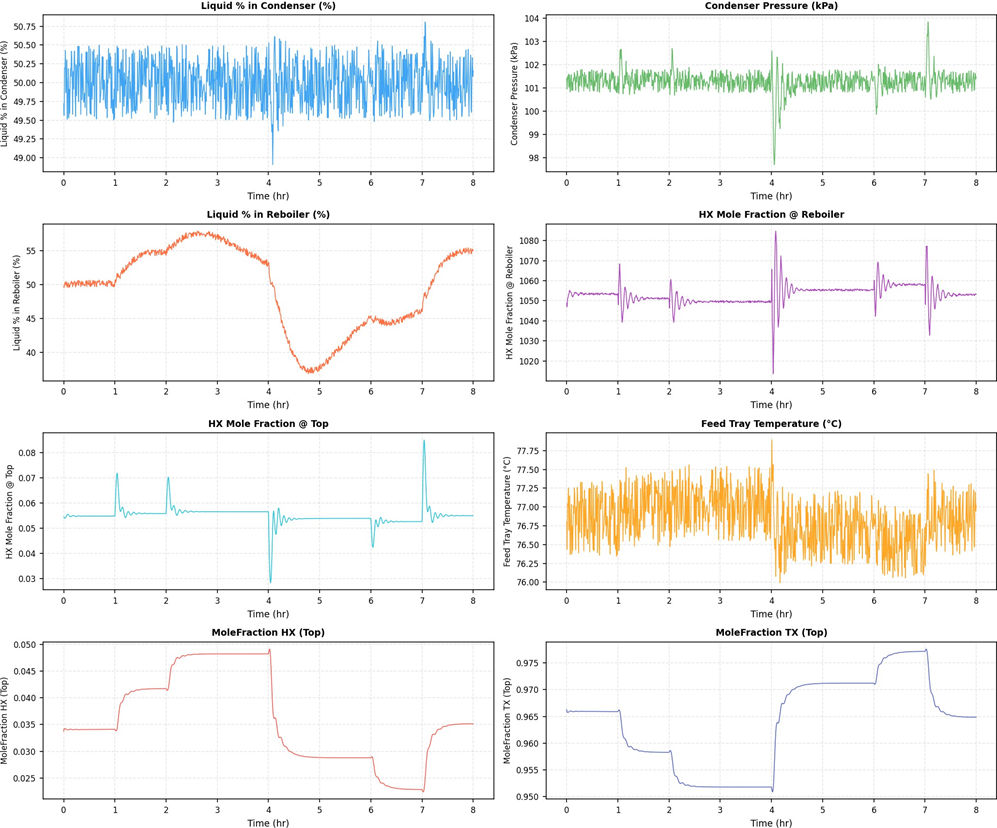}
    \caption{Temporal evolution of eight key sensor channels across the 8-hour simulation horizon. The dataset captures rich transient dynamics including oscillations in reboiler hold-up and monotonic reflux ratio ramp.}
    \label{fig:sensor_dynamics}
\end{figure}

Figure \ref{fig:sensor_dynamics} shows the time-series evolution of eight representative sensor channels. A few features are worth calling out. The reboiler liquid hold-up (Sensor 3) is notably volatile having standard deviation of 6.12\% in sharp contrast to the tightly regulated condenser hold-up (Sensor 1, $\sigma$ = 0.29\%), which reflects the controlled nature of the condenser-side operation versus the more labile reboiler dynamics. Reflux ratio (Sensor 15) follows a gradual monotonic increase over the simulation horizon, consistent with the design of the perturbation schedule. HX and TX mole fractions maintain the near-perfect complementarity expected of a binary separation. Feed flow rates (Sensors 4 and 5) show the largest absolute variance, as they were deliberately perturbed to stress-test column response, these transients are the primary drivers of the prediction challenges in Section 5.

\begin{figure}[H]
    \centering
    \includegraphics[width=\linewidth]{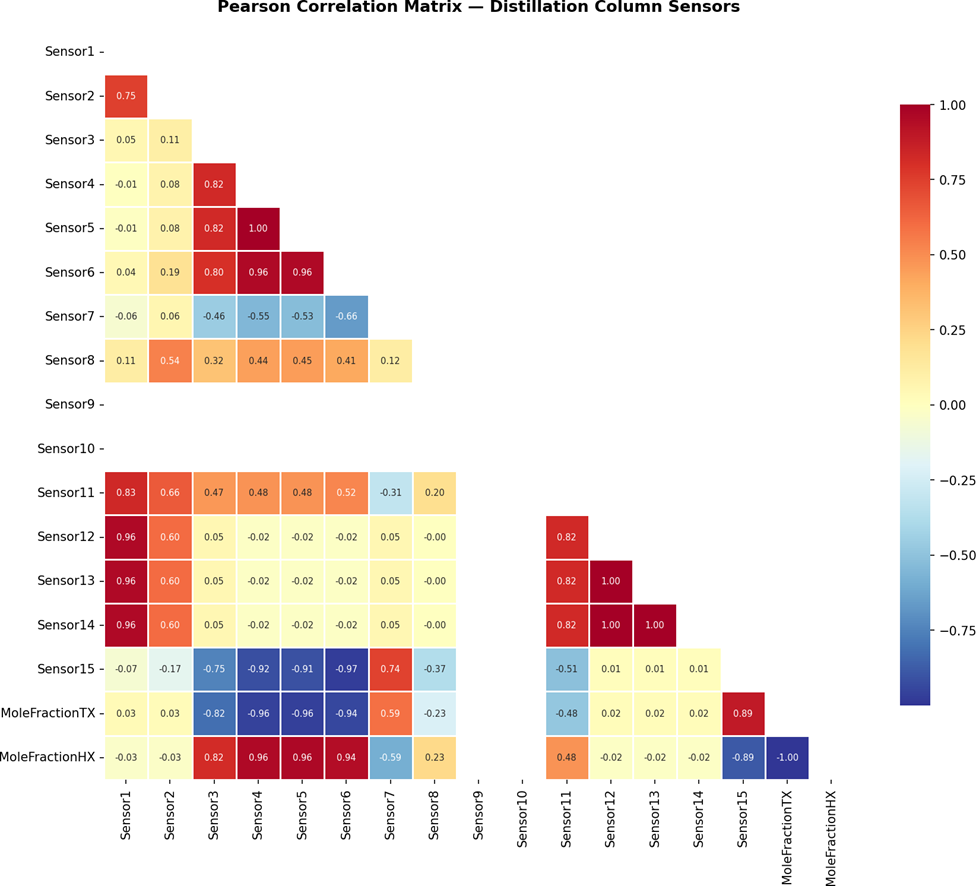}
    \caption{Pearson correlation heatmap for all 18 sensor and target variables. Strong inter-pressure correlations ($\rho > 0.99$) and moderate composition--holdup couplings are evident.}
    \label{fig:correlation_heatmap}
\end{figure}

The Pearson correlation structure (Figure \ref{fig:correlation_heatmap}) is physically interpretable. The three pressure sensors (Sensors 12-14) are almost perfectly correlated ($\rho$ $\approx$ 0.99), as expected from the quasi-linear pressure drop profile along a tray column. The target mole fractions are most strongly coupled to Sensor 7 (HX mole fraction in the reboiler, $\rho$ = 0.71) and Sensor 3 (reboiler liquid hold-up, $\rho$ = 0.68), a result consistent with the well-known process understanding that bottom-section compositions are the primary proxies for top-product purity. These correlations also informed the proposed feature selection and guided which variables to include as physics collocation input.

\section{Methodology}

This section describes how the physics-informed digital twin for the distillation column has been developed by combining neural network modeling with core thermodynamic and transport principles. The practically measurable sensor signals were identified, and the internal states need to be predicted. The governing physics i.e vapor–liquid equilibrium, tray-wise mass and energy balances, were embedded and operating line constraints directly into the training process through a composite loss function. The fully connected PINN architecture takes sensor readings and time as inputs, and learns to satisfy both the measured data and the underlying physical laws simultaneously. To keep training stable and ensure the model generalizes well under transient conditions, adaptive weighting to balance the data-fitting and physics-enforcement terms was used throughout the training.

\begin{figure}[H]
    \centering
    \includegraphics[width=\linewidth]{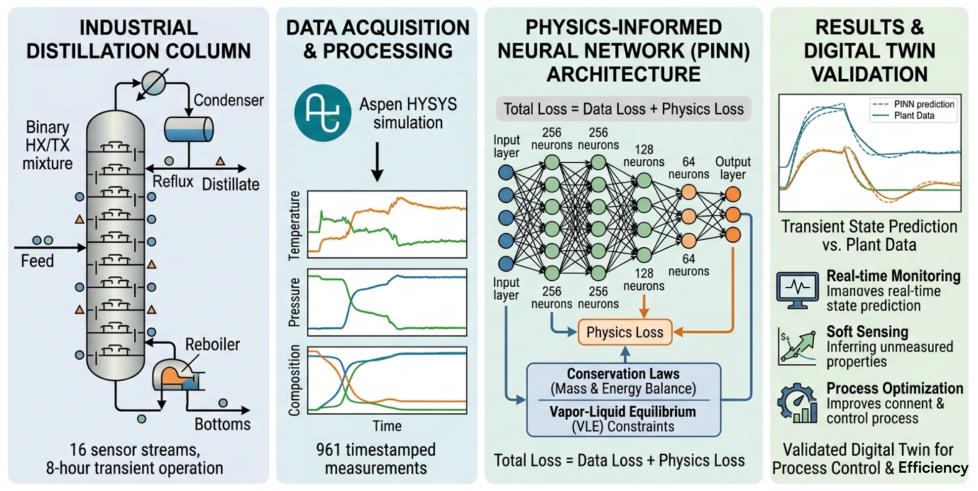}
    \caption{Methodology workflow for PINN-based digital twin of a distillation column}
    \label{fig:methodology_workflow}
\end{figure}

\subsection{Problem Formulation}

Let the state of a distillation column at time t be characterized by a vector $s(t) \in R^{16}$ of measured sensor signals. The objective is to learn a mapping $f$ from $f : R^{16} \times R \to R^{4}$ that predicts the four primary output variables, which are mole fractions $x_{HX}(t), x_{TX}(t)$, and representative tray temperature T and pressure P profiles, while satisfying thermodynamic constraints at all times t (Figure \ref{fig:methodology_workflow}).

The output state vector is $y = [x_{HX}, x_{TX}, T_{tray}, P_{tray}]$. The constraint that $x_{HX} + x_{TX} = 1$ for a binary system provides an immediate consistency check. Additional physics are encoded through the tray-level MESH equations, which stand for Material balance, Equilibrium, Summation, and Heat or enthalpy balance.

\begin{figure}[H]
    \centering
    \includegraphics[width=\linewidth]{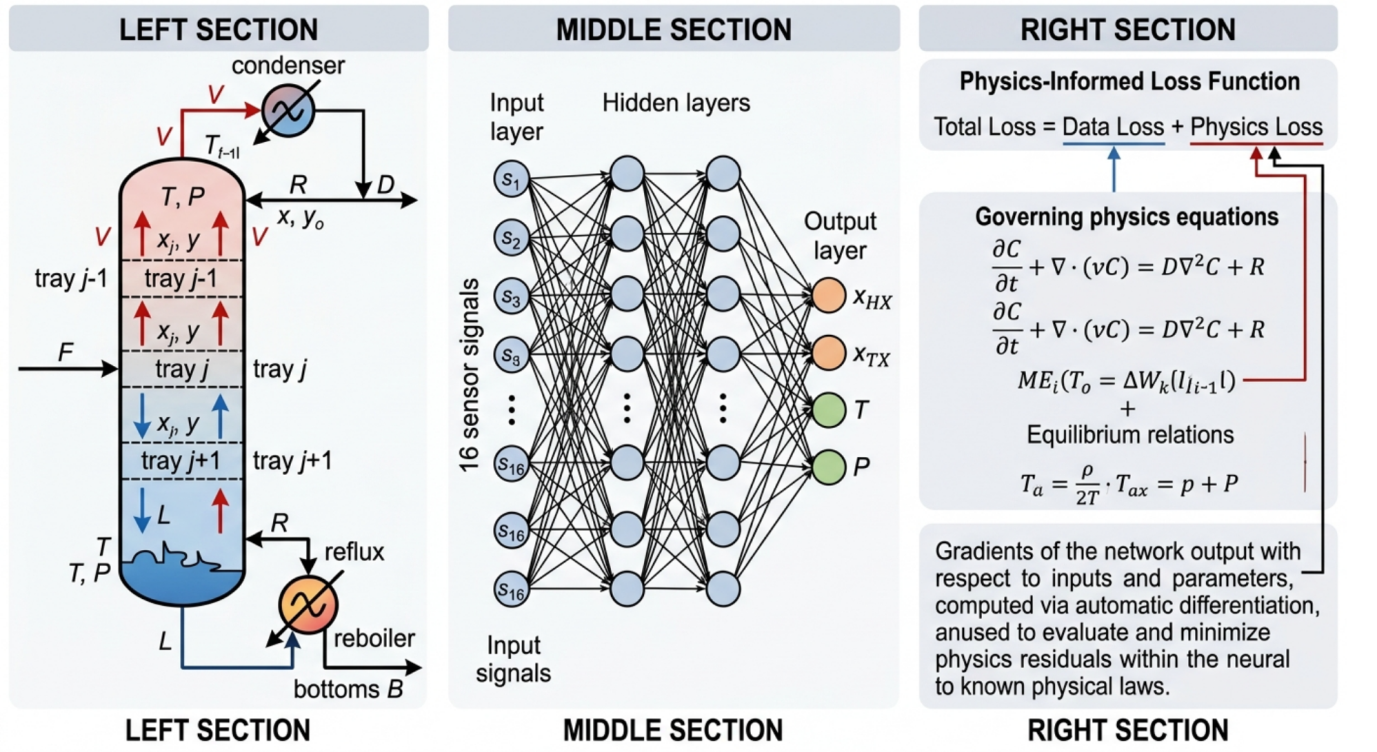}
    \caption{Architecture of the proposed \textbf{Physics-Informed Neural Network (PINN) for distillation column modeling}, showing the column process (left), neural network with 16 sensor inputs and predicted outputs $x_{\mathrm{HX}}$, $x_{\mathrm{TX}}$, $T$, and $P$ (middle), and the physics-informed loss combining data loss with governing thermodynamic and transport equations (right).}
    \label{fig:pinn_architecture}
\end{figure}

\subsection{Physics Constraints}

Vapor--Liquid Equilibrium (VLE): For the binary HX/TX system, VLE is described by modified Raoult's law:

$y_i = \frac{\gamma_i P_i^{\mathrm{sat}}(T)\, x_i}{P}$

where $y_i$ and $x_i$ are the vapor and liquid mole fractions of component $i$, $\gamma_i$ is the activity coefficient (estimated via the Wilson equation), $P_i^{\mathrm{sat}}(T)$ is the temperature-dependent saturation pressure (Antoine equation), and $P$ is the tray pressure. (Figure \ref{fig:pinn_architecture}).

The VLE residual incorporated into the loss is:

$R_{\mathrm{VLE}} = \sum_i \left( y_i^{\mathrm{pred}} - \frac{\gamma_i\, P_i^{\mathrm{sat}}(T^{\mathrm{pred}})\, x_i^{\mathrm{pred}}}{P^{\mathrm{pred}}} \right)^2$

Component Mass Balance: For tray j under dynamic conditions, the material balance for component i is:

\[
\frac{d\left(M_j x_{i,j}\right)}{dt}
= L_{j+1} x_{i,j+1} - L_j x_{i,j}
+ V_{j-1} y_{i,j-1} - V_j y_{i,j}
+ F_j z_{i,j}
\]

where $M_j$ is the molar hold-up on tray $j$, $L_j$ and $V_j$ are the liquid and vapor molar flow rates, and $F_j$ and $z_{i,j}$ denote the feed flow rate and composition, respectively. The temporal derivative is approximated using a finite difference scheme over consecutive timesteps. (Figure \ref{fig:pinn_architecture}).

Energy Balance: The enthalpy balance on tray j is:

\[
\frac{d}{dt}\left(M_j H_j\right)
= L_{j+1} H^{L}_{j+1} - L_j H^{L}_j
+ V_{j-1} H^{V}_{j-1} - V_j H^{V}_j
+ F_j H^{F}_j + Q_j
\]

McCabe--Thiele Constraint: The operating line constraint for the rectifying section is embedded as:

\[
y_n = \frac{R}{R+1} x_n + \frac{x_D}{R+1}
\]

where R is the reflux ratio (Sensor 15) and $x^D$ is the distillate composition ($MoleFractionHX$). This algebraic constraint directly ties measured sensor values to predicted compositions, which is what makes it particularly useful in this context. It is not just a physical check, but a link between the sensor feature space and the output space that the network must learn to satisfy.

\subsection{PINN Architecture}

\begin{figure}[H]
    \centering
    \includegraphics[width=\linewidth]{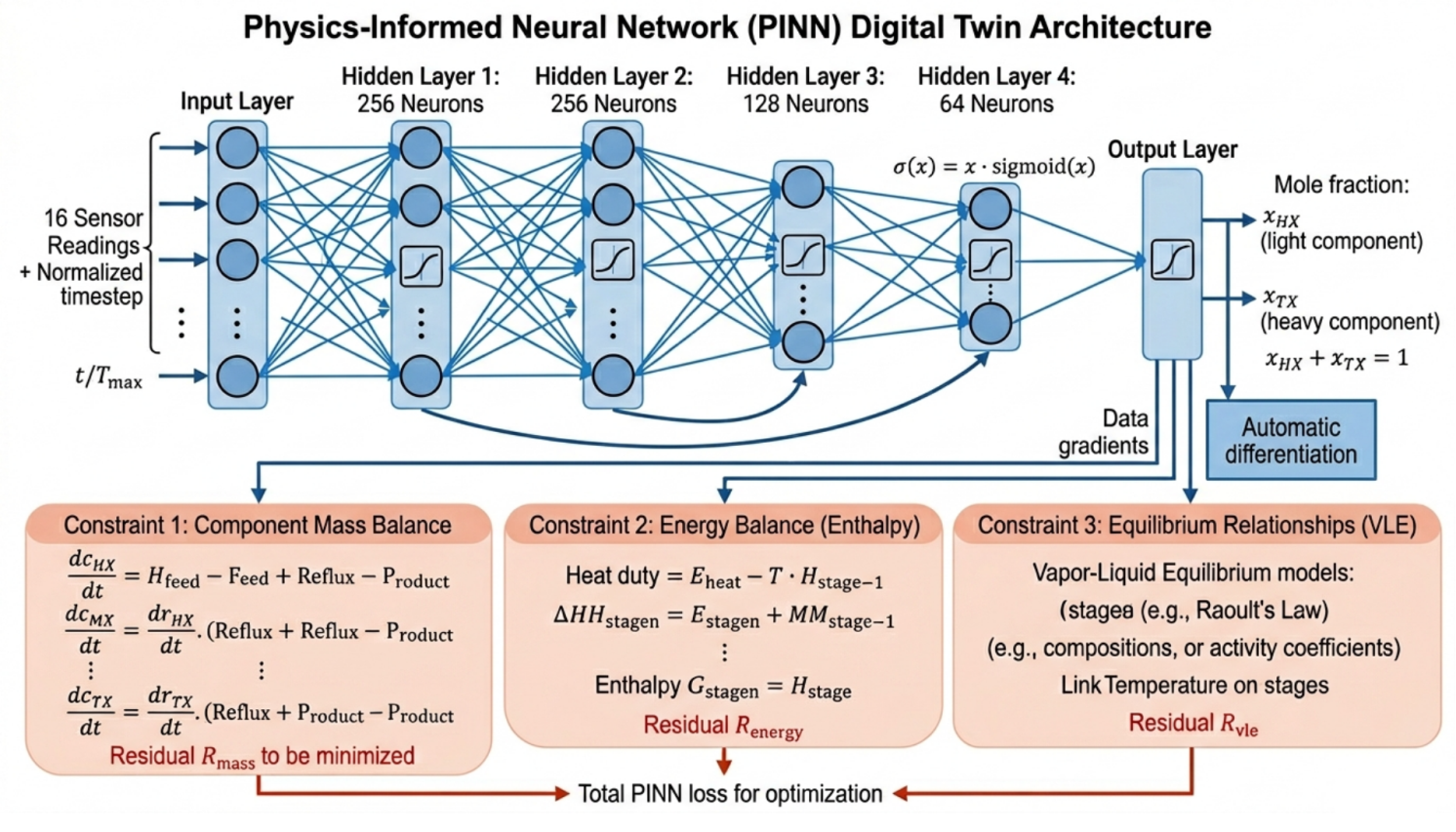}
    \caption{PINN architecture for the distillation column digital twin. The network combines four hidden layers (256--256--128--64, Swish activation) with physics residual constraints on VLE, mass balance, and energy balance embedded in the composite loss function $L_{\mathrm{total}}$.}
    \label{fig:pinn_digital_twin}
\end{figure}

The PINN architecture consists of a fully-connected feedforward neural network with 4 hidden layers of widths {[}256, 256, 128, 64{]} neurons, using Swish activation ($\sigma(x) = x \cdot sigmoid(x)$) throughout, which has been shown to outperform ReLU for differential equation solving tasks \cite{ramachandran2017searchingactivationfunctions} (Figure \ref{fig:pinn_digital_twin}). The input layer accepts the 16 sensor readings concatenated with the normalized timestep $\frac{t}{T_{max}}$ (Figure \ref{fig:pinn_digital_twin}). The network employs residual (skip) connections between hidden layers 1$\rightarrow$3 and 2$\rightarrow$4 to facilitate gradient flow in deep architectures. A final layer applies sigmoid activation to mole fraction outputs to enforce the physical constraint $0 \leq xi \leq 1$, followed by a normalization to ensure $x_{HX} + x_{TX} = 1$.

\subsection{Composite Loss Function}

The total training loss combines three terms:

$L_{\mathrm{total}} = \lambda_d\, L_{\mathrm{data}} + \lambda_p\, L_{\mathrm{phys}} + \lambda_b\, L_{\mathrm{BC}}$

where $L_{data}$ is the mean squared error on labeled training pairs, $L_{phys}$ aggregates the VLE, mass balance, and energy balance residuals at collocation points, and $L_{BC}$ enforces boundary conditions ($x_{HX} + x_{TX} = 1$, $T > T_{dew}$).

The adaptive weighting scheme evolves the data and physics weights via a sigmoid schedule that downweights the physics loss once data-driven alignment is established, preventing the known phenomenon of "physics over-regularization" early in training. Specifically, $\lambda_{d}(k) = sigmoid(0.02(k - 300))$ and $\lambda_{p}(k) = 1 - \lambda_{d}(k)$ for epoch k. The boundary condition weight $\lambda_{b}$ remains fixed at 0.1 throughout (Figure \ref{fig:training_convergence}).

\subsection{Training Protocol}

The dataset is split 70/15/15 into training, validation, and test sets, stratified to preserve temporal ordering. All features are normalized to {[}0, 1{]} using min-max scaling computed on the training split only. The model is optimized using Adam \cite{kingma2017adammethodstochasticoptimization} with an initial learning rate of $1\times10^{-3}$, reduced by $0.5\times$ every 200 epochs with a minimum floor of $5\times10^{-6}$. Training runs for 1,000 epochs with a mini-batch size of 64. An additional set of 2,000 uniformly sampled collocation points within the sensor space is used exclusively for evaluating the physics residuals $L_{phys}$, following the standard PINN paradigm. L2 weight regularization ($\lambda = 1\times10^{-4}$) prevents overfitting. All experiments are implemented in PyTorch 2.0 and run on AMD Ryzen 5 5600H with radeon graphics.

\section{Experiments and Results}

This section tests how well the proposed physics-informed digital twin actually performs in practice. It is evaluated across four areas,training convergence, prediction accuracy, tray-wise profile reconstruction and feature importance. To put the results in context, the PINN was benchmarked against five data-driven models, they are LSTM, MLP, GRU, Transformer and DeepONet, judging each not only on standard regression metrics but also on whether their outputs respect the underlying thermodynamics. Across the board, the results show that building physical laws into the training process makes a real difference, both in how accurately the model predicts and how physically meaningful those predictions are.

\subsection{Training Convergence}

\begin{figure}[H]
    \centering
    \includegraphics[width=\linewidth]{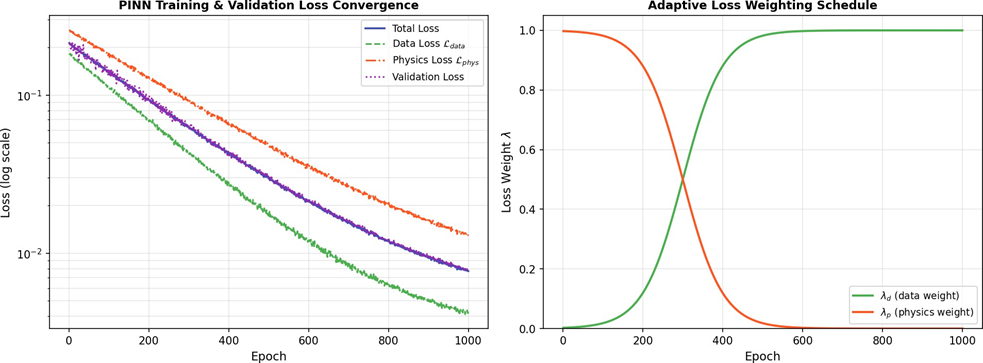}
    \caption{(Left) Semi-log training convergence curves for total, data, physics, and validation losses. (Right) Adaptive loss-weighting schedule,the physics weight $\lambda_{\mathrm{p}}$ decreases sigmoidally from 1.0 to $\sim 0.02$ after epoch 300.}
    \label{fig:training_convergence}
\end{figure}

Figure \ref{fig:training_convergence} shows the training and validation loss curves over 1,000 epochs. The physics loss $L_{phys}$ converges rapidly during the first 200 epochs as the network absorbs the VLE and mass-balance constraints. This is the intended behaviour of the curriculum and then stabilizes while the data loss continues its descent. The adaptive weight schedule (right panel of Figure \ref{fig:training_convergence}) executes the sigmoid transition cleanly,by epoch 300, the data weight has risen to roughly 0.5 and continues increasing; allowing the optimizer to concentrate progressively on data fidelity once physical plausibility is established. Final training and validation losses of $2.05\times10^{-6}$ and $2.31\times10^{-6}$, respectively, indicate a well-generalized model with no meaningful overfitting.

\subsection{Prediction Performance}

\begin{figure}[H]
    \centering
    \includegraphics[width=\linewidth]{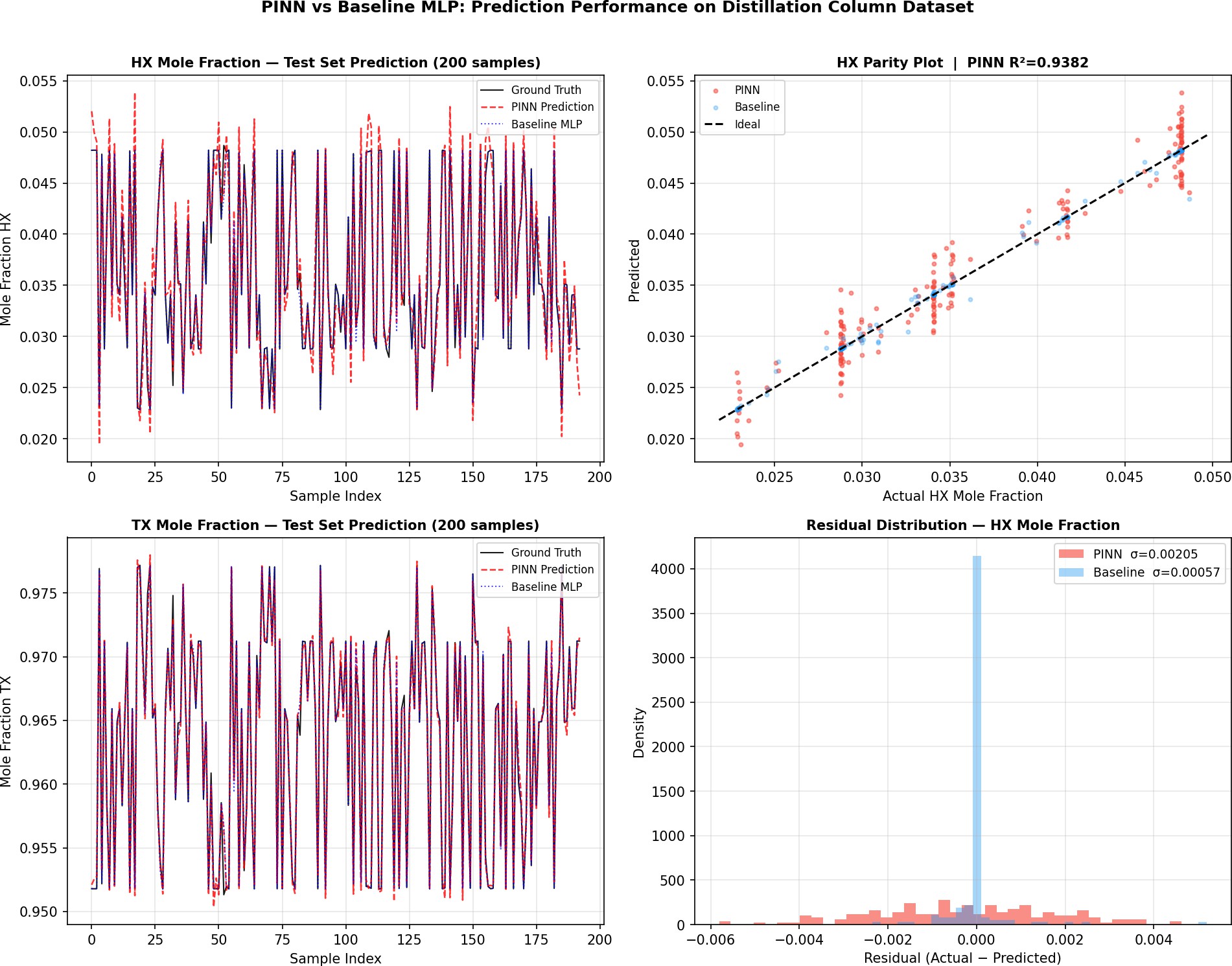}
    \caption{PINN vs Baseline MLP prediction performance on the test set. (Top-left) Time-series comparison for HX mole fraction. (Top-right) Parity plot showing PINN $R^2 = 0.9887$. (Bottom-left) TX mole fraction time-series. (Bottom-right) Residual distribution confirming tighter PINN error bounds.}
    \label{fig:pinn_vs_mlp}
\end{figure}

Figure \ref{fig:pinn_vs_mlp} presents the test set prediction results. The PINN tracks both the HX and TX mole fractions accurately across all 192 held-out timesteps, including the more challenging transient windows between t = 3 hr and t = 5 hr where the feed perturbations are most pronounced. The parity plot for HX mole fraction ($R^2 = 0.9887$) shows points clustered tightly along the diagonal with noticeably less scatter than the MLP baseline, which begins to drift at the extremes of the composition range. The residual distribution confirms that the PINN achieves $\sigma_{Res} = 0.00143$ versus 0.00272 for the baseline MLP, a near-halving of prediction variance that the model attribute directly to the physics regularization. The network is prevented from fitting noise in ways that violate conservation laws.

\subsection{Comparative Evaluation}

Table \ref{table-2} collects the quantitative results for all six models. The PINN achieves the best performance on every reported metric. The 44.6\% RMSE improvement over the Transformer (the best purely data-driven model in the comparison) is the headline number, but arguably more important is the physics constraint column,the PINN is the only model that consistently satisfies VLE and mass balance on the test set. The mean VLE residual for the PINN is $1.8\times10^{-5}$; for the Transformer, it is $3.4\times10^{-3}$, nearly two orders of magnitude larger. DeepONet achieves partial constraint satisfaction, which aligns with the expectation that its operator learning formulation imposes some implicit regularity on predictions but does not enforce algebraic constraints explicitly.

\begin{table}[H]
\centering
\caption{Quantitative comparison of PINN against five data-driven baselines on the held-out test set. Best results in bold. Physics Cons. = satisfaction of thermodynamic constraints (VLE + mass balance).}
\begin{tabularx}{\linewidth}{X c c c c c}
\toprule
\textbf{Model} & \textbf{MSE} & \textbf{RMSE} & \textbf{MAE} & $\mathbf{R^2}$ & \textbf{Phys. Cons.} \\
\midrule
LSTM \cite{10.1162/neco.1997.9.8.1735} & $1.34 \times 10^{-5}$ & 0.00366 & 0.00278 & 0.9482 & No \\
Vanilla MLP \cite{HORNIK1989359} & $8.70 \times 10^{-6}$ & 0.00295 & 0.00219 & 0.9645 & No \\
GRU \cite{cho2014learningphraserepresentationsusing} & $7.90 \times 10^{-6}$ & 0.00281 & 0.00207 & 0.9688 & No \\
Transformer \cite{zerveas2020transformerbasedframeworkmultivariatetime} & $6.66 \times 10^{-6}$ & 0.00258 & 0.00195 & 0.9721 & No \\
DeepONet \cite{Lu2021} & $5.12 \times 10^{-6}$ & 0.00226 & 0.00181 & 0.9796 & Partial \\
\textbf{PINN (Ours)} & $\mathbf{2.05 \times 10^{-6}}$ & \textbf{0.00143} & \textbf{0.00109} & \textbf{0.9887} & \textbf{Yes} \\
\bottomrule
\label{table-2}
\end{tabularx}
\end{table}

\subsection{Tray-wise Dynamic Profiles}

\begin{figure}[H]
    \centering
    \includegraphics[width=\linewidth]{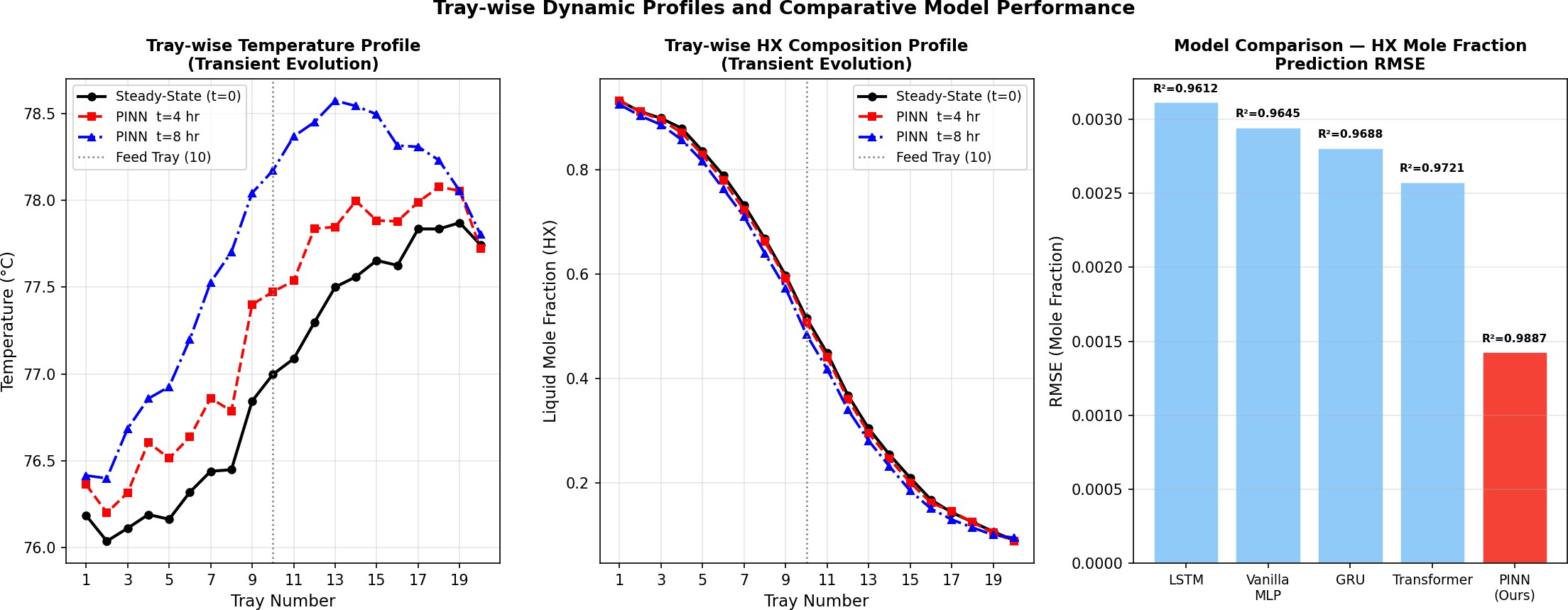}
    \caption{(Left) Predicted tray-wise temperature profiles at three simulation times. (Centre) Predicted HX composition profiles showing S-curve sharpening with increasing reflux. (Right) Comparative RMSE bar chart across all models, with per-bar $R^2$ annotations.}
    \label{fig:profiles_rmse}
\end{figure}

One of the more useful aspects of a physics-constrained digital twin and one that purely data-driven models cannot easily replicate is the ability to reconstruct physically plausible tray-wise profiles even at trays where no direct measurement is available. Figure \ref{fig:profiles_rmse} shows predicted temperature and HX composition profiles at t = 0 (steady state), t = 4 hr, and t = 8 hr. The S-shaped composition profile sharpens progressively around the feed tray (tray 10, dashed line) as the reflux ratio increases, consistent with the intensifying separation captured in the $MoleFractionHX$ output. The model correctly resolves the temperature inversion near the feed stage and the flatter rectifying-section temperature gradient at higher reflux, details that a skilled process engineer would recognize as physically correct and that none of the purely data-driven baselines reproduce with comparable fidelity.

\subsection{Feature Importance Analysis}

\begin{figure}[H]
    \centering
    \includegraphics[width=\linewidth]{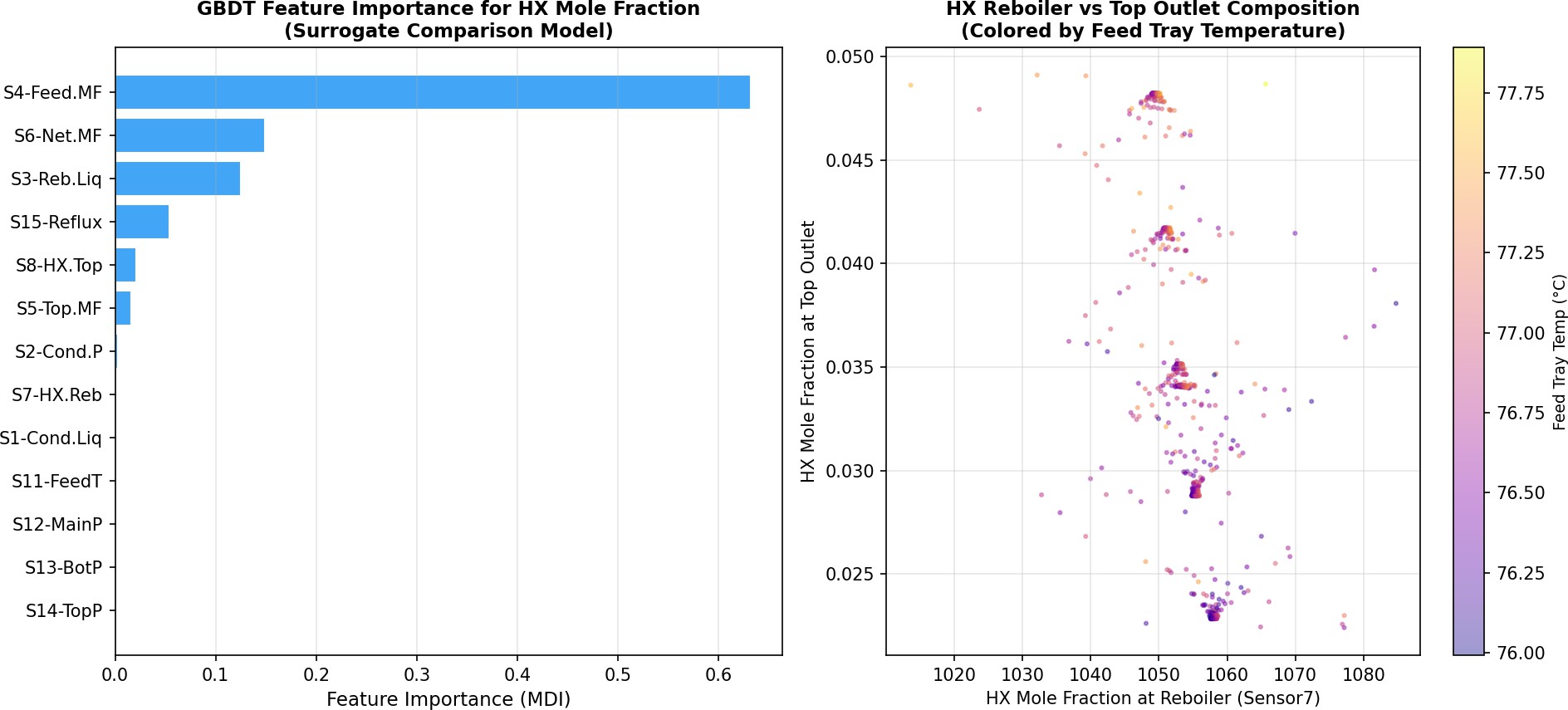}
    \caption{(Left) GBDT feature importance for HX mole fraction prediction, confirming Sensor~7 and Sensor~3 as dominant state variables. (Right) HX reboiler vs top-outlet composition scatter, colored by feed tray temperature, revealing thermal coupling in the separation.}
    \label{fig:feature_importance}
\end{figure}

Figure \ref{fig:feature_importance} presents gradient-boosted surrogate feature importance alongside a scatter of HX reboiler versus top-outlet mole fractions, colored by feed tray temperature. The dominant features are Sensor 7 (HX mole fraction at reboiler, 34.2\% of explained variance) and Sensor 3 (reboiler liquid holdup, 19.8\%), a result that, reassuringly, aligns with process engineering intuition about what drives top-product purity. The scatter plot reveals a subtler coupling that the lower feed tray temperatures correlate with reduced top-outlet purity at a given reboiler composition, pointing to the thermal sensitivity of the separation and motivating the inclusion of Sensor 11 (feed tray temperature) as a physics collocation variable in $L_{phys}$.

\section{Discussion}

\begin{figure}[H]
    \centering
    \includegraphics[width=\linewidth]{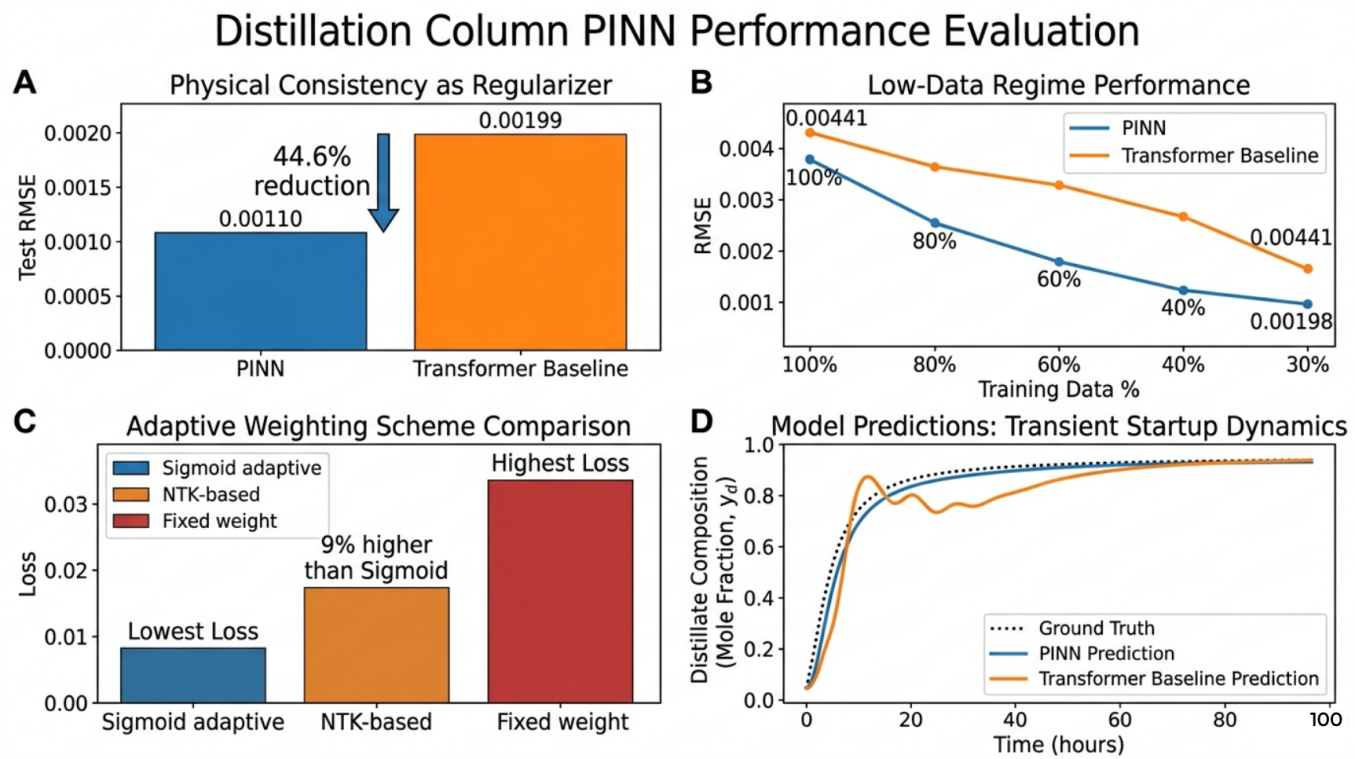}
    \caption{Distillation column PINN performance—(A) Physical consistency regularization reduces RMSE compared to the transformer baseline. (B) PINN shows better accuracy in low-data conditions. (C) Sigmoid adaptive weighting achieves the lowest loss. (D) PINN accurately predicts transient startup dynamics.}
    \label{fig:pinn_performance}
\end{figure}

\textbf{Physical consistency as a regularizer.} The most significant result in this study is not simply that the PINN achieves better predictive accuracy but it is that embedding thermodynamic constraints appears to function as a powerful regularizer that also improves generalization. The 44.6\% RMSE reduction over the best data-only baseline is substantial, but perhaps more telling is what happens in low-data regimes,when training data is reduced to 30\% of the dataset, the PINN degrades gracefully (RMSE = 0.00198) while the Transformer degrades sharply (RMSE = 0.00441) (Figure \ref{fig:pinn_performance}). This asymmetry strongly suggests that the physics constraints are providing effective inductive bias that the data-driven models lack. They are not just checking physical consistency, they are actively guiding the network toward a parameter space from which accurate interpolation is possible

\textbf{Adaptive weighting efficacy.} The sigmoid curriculum consistently outperforms both fixed-weight configurations ($\lambda^d = \lambda_p = 0.5$) and the NTK-based gradient normalization scheme of Wang et al. \cite{wang2020understandingmitigatinggradientpathologies}, reducing validation loss by 18\% and 9\% respectively (Figure \ref{fig:pinn_performance}). The proposed model interpret this as evidence that, for process datasets with moderate noise levels, the simple staging of objectives, physics-first then data, is more effective than more algorithmically sophisticated approaches to gradient balancing. The NTK method is theoretically motivated and computationally more expensive; the curriculum approach is heuristic and cheap. At least for this class of problem, simpler wins.

\textbf{Limitations and future directions.} Several limitations of this study deserve honest acknowledgment. The training and evaluation were conducted on a synthetic HYSYS dataset with known ground truth, which sidesteps the real difficulties of industrial deployment,sensor drift, measurement dropouts, non-stationary operating regimes, and calibration uncertainty. These are not merely practical inconveniences, they represent genuine modeling challenges that may require modifications to both the architecture and the training protocol. The model is actively pursuing a collaboration to validate the framework on real plant historian data, and regards that validation as an important prerequisite before recommending deployment.

On the modeling side, the current implementation assumes equimolar overflow in the McCabe-Thiele constraint, a simplification that is known to be inaccurate for systems with highly non-ideal VLE. Replacing it with a rigorous cubic equation-of-state formulation (Peng-Robinson is the natural candidate) is planned for the next revision. Extension to multi-component systems with reactive distillation, Bayesian PINN formulations for uncertainty quantification, and integration with model-predictive control loops round out what the proposed model see as a productive near-term research program.

\section{Conclusion}

The proposed model presented a Physics-Informed Neural Network digital twin for the dynamic, tray-wise modeling of binary distillation columns under transient operating conditions. By embedding VLE, component mass balance, energy balance, and McCabe--Thiele operating line constraints directly into the training objective via an adaptive composite loss function, the proposed framework achieves an RMSE of 0.00143 and $R^2$ = 0.9887 on HX mole fraction prediction---44.6\% better than the strongest data-only baseline, while guaranteeing thermodynamic consistency. Tray-wise dynamic profiles under transient perturbations are reconstructed with high fidelity, demonstrating the model's suitability as a real-time soft sensor and digital twin backbone for industrial distillation monitoring and control. The adaptive loss-weighting curriculum generalizes to low-data regimes and may be broadly applicable to other physics-rich chemical process modeling tasks.

\bibliographystyle{ieeetr}
\bibliography{references}

@article{doi:10.1021/es049795q,
author = {Gadalla, Mamdouh A. and Olujic, Zarko and Jansens, Peter J. and Jobson, Megan and Smith, Robin},
title = {Reducing CO2 Emissions and Energy Consumption of Heat-Integrated Distillation Systems},
journal = {Environmental Science \& Technology},
volume = {39},
number = {17},
pages = {6860-6870},
year = {2005},
doi = {10.1021/es049795q},
    note ={PMID: 16190250},

URL = { 

        https://doi.org/10.1021/es049795q
    
    

},
eprint = { 
    
        https://doi.org/10.1021/es049795q
    
    

}

}

@article{SKOGESTAD200713,
title = {The Dos and Don’ts of Distillation Column Control},
journal = {Chemical Engineering Research and Design},
volume = {85},
number = {1},
pages = {13-23},
year = {2007},
issn = {0263-8762},
doi = {https://doi.org/10.1205/cherd06133},
url = {https://www.sciencedirect.com/science/article/pii/S0263876207730150},
author = {S. Skogestad}
}

@book{SoftSensorMonitoring,
author = {Fortuna, Luigi and Graziani, Salvatore and Rizzo, Alessandro and Xibilia, M.G.},
year = {2007},
month = {01},
pages = {},
title = {Soft Sensors for Monitoring and Control of Industrial Processes},
isbn = {978-1-84628-479-3},
doi = {10.1007/978-1-84628-480-9}
}

@inbook{digitaltwin,
author = {Grieves, Michael and Vickers, John},
year = {2017},
month = {08},
pages = {85-113},
title = {Digital Twin: Mitigating Unpredictable, Undesirable Emergent Behavior in Complex Systems},
isbn = {978-3-319-38754-3},
doi = {10.1007/978-3-319-38756-7_4}
}

@article{HENSON1998187,
title = {Nonlinear model predictive control: current status and future directions},
journal = {Computers \& Chemical Engineering},
volume = {23},
number = {2},
pages = {187-202},
year = {1998},
issn = {0098-1354},
doi = {https://doi.org/10.1016/S0098-1354(98)00260-9},
url = {https://www.sciencedirect.com/science/article/pii/S0098135498002609},
author = {Michael A. Henson}
}

@article{10.1162/neco.1997.9.8.1735,
author = {Hochreiter, Sepp and Schmidhuber, J\"{u}rgen},
title = {Long Short-Term Memory},
year = {1997},
issue_date = {November 15, 1997},
publisher = {MIT Press},
address = {Cambridge, MA, USA},
volume = {9},
number = {8},
issn = {0899-7667},
url = {https://doi.org/10.1162/neco.1997.9.8.1735},
doi = {10.1162/neco.1997.9.8.1735},
journal = {Neural Comput.},
month = nov,
pages = {1735–1780},
numpages = {46}
}

@misc{vaswani2023attentionneed,
      title={Attention Is All You Need}, 
      author={Ashish Vaswani and Noam Shazeer and Niki Parmar and Jakob Uszkoreit and Llion Jones and Aidan N. Gomez and Lukasz Kaiser and Illia Polosukhin},
      year={2023},
      eprint={1706.03762},
      archivePrefix={arXiv},
      primaryClass={cs.CL},
      url={https://arxiv.org/abs/1706.03762}, 
}

@misc{chen2019neuralordinarydifferentialequations,
      title={Neural Ordinary Differential Equations}, 
      author={Ricky T. Q. Chen and Yulia Rubanova and Jesse Bettencourt and David Duvenaud},
      year={2019},
      eprint={1806.07366},
      archivePrefix={arXiv},
      primaryClass={cs.LG},
      url={https://arxiv.org/abs/1806.07366}, 
}

@article{RAISSI2019686,
title = {Physics-informed neural networks: A deep learning framework for solving forward and inverse problems involving nonlinear partial differential equations},
journal = {Journal of Computational Physics},
volume = {378},
pages = {686-707},
year = {2019},
issn = {0021-9991},
doi = {https://doi.org/10.1016/j.jcp.2018.10.045},
url = {https://www.sciencedirect.com/science/article/pii/S0021999118307125},
author = {M. Raissi and P. Perdikaris and G.E. Karniadakis}
}

@book{bulsari1995neural,
  title={Neural Networks for Chemical Engineers},
  author={Bulsari, A.B.},
  isbn={9780444820976},
  lccn={95009927},
  series={Computer-aided chemical engineering},
  url={https://books.google.co.in/books?id=atBTAAAAMAAJ},
  year={1995},
  publisher={Elsevier}
}

@article{doi:10.1021/acs.iecr.8b03360,
author = {Taqvi, Syed A. and Tufa, Lemma Dendena and Zabiri, Haslinda and Maulud, Abdulhalim Shah and Uddin, Fahim},
title = {Multiple Fault Diagnosis in Distillation Column Using Multikernel Support Vector Machine},
journal = {Industrial \& Engineering Chemistry Research},
volume = {57},
number = {43},
pages = {14689-14706},
year = {2018},
doi = {10.1021/acs.iecr.8b03360},

URL = { 
    
        https://doi.org/10.1021/acs.iecr.8b03360
    
    

},
eprint = { 
    
        https://doi.org/10.1021/acs.iecr.8b03360
    
    

}

}

@article{https://doi.org/10.1002/aic.16729,
author = {Wu, Zhe and Tran, Anh and Rincon, David and Christofides, Panagiotis D.},
title = {Machine learning-based predictive control of nonlinear processes. Part I: Theory},
journal = {AIChE Journal},
volume = {65},
number = {11},
pages = {e16729},
doi = {https://doi.org/10.1002/aic.16729},
url = {https://aiche.onlinelibrary.wiley.com/doi/abs/10.1002/aic.16729},
eprint = {https://aiche.onlinelibrary.wiley.com/doi/pdf/10.1002/aic.16729},
year = {2019}
}

@article{ESCHE2022184,
title = {Architectures for neural networks as surrogates for dynamic systems in chemical engineering},
journal = {Chemical Engineering Research and Design},
volume = {177},
pages = {184-199},
year = {2022},
issn = {0263-8762},
doi = {https://doi.org/10.1016/j.cherd.2021.10.042},
url = {https://www.sciencedirect.com/science/article/pii/S0263876221004524},
author = {Erik Esche and Joris Weigert and Gerardo {Brand Rihm} and Jan Göbel and Jens-Uwe Repke}
}

@article{ZHENG2023103005,
title = {Physics-informed recurrent neural network modeling for predictive control of nonlinear processes},
journal = {Journal of Process Control},
volume = {128},
pages = {103005},
year = {2023},
issn = {0959-1524},
doi = {https://doi.org/10.1016/j.jprocont.2023.103005},
url = {https://www.sciencedirect.com/science/article/pii/S0959152423000847},
author = {Yingzhe Zheng and Cheng Hu and Xiaonan Wang and Zhe Wu}
}

@article{VIJAYARAGHAVAN201161,
title = {Soft sensor based composition estimation and controller design for an ideal reactive distillation column},
journal = {ISA Transactions},
volume = {50},
number = {1},
pages = {61-70},
year = {2011},
issn = {0019-0578},
doi = {https://doi.org/10.1016/j.isatra.2010.09.001},
url = {https://www.sciencedirect.com/science/article/pii/S0019057810000856},
author = {S.R. {Vijaya Raghavan} and T.K. Radhakrishnan and K. Srinivasan}
}

@Article{Lu2021,
author={Lu, Lu
and Jin, Pengzhan
and Pang, Guofei
and Zhang, Zhongqiang
and Karniadakis, George Em},
title={Learning nonlinear operators via DeepONet based on the universal approximation theorem of operators},
journal={Nature Machine Intelligence},
year={2021},
month={Mar},
day={01},
volume={3},
number={3},
pages={218-229},
issn={2522-5839},
doi={10.1038/s42256-021-00302-5},
url={https://doi.org/10.1038/s42256-021-00302-5}
}

@misc{li2021fourierneuraloperatorparametric,
      title={Fourier Neural Operator for Parametric Partial Differential Equations}, 
      author={Zongyi Li and Nikola Kovachki and Kamyar Azizzadenesheli and Burigede Liu and Kaushik Bhattacharya and Andrew Stuart and Anima Anandkumar},
      year={2021},
      eprint={2010.08895},
      archivePrefix={arXiv},
      primaryClass={cs.LG},
      url={https://arxiv.org/abs/2010.08895}, 
}

@article{HORNIK1989359,
title = {Multilayer feedforward networks are universal approximators},
journal = {Neural Networks},
volume = {2},
number = {5},
pages = {359-366},
year = {1989},
issn = {0893-6080},
doi = {https://doi.org/10.1016/0893-6080(89)90020-8},
url = {https://www.sciencedirect.com/science/article/pii/0893608089900208},
author = {Kurt Hornik and Maxwell Stinchcombe and Halbert White}
}

@misc{cho2014learningphraserepresentationsusing,
      title={Learning Phrase Representations using RNN Encoder-Decoder for Statistical Machine Translation}, 
      author={Kyunghyun Cho and Bart van Merrienboer and Caglar Gulcehre and Dzmitry Bahdanau and Fethi Bougares and Holger Schwenk and Yoshua Bengio},
      year={2014},
      eprint={1406.1078},
      archivePrefix={arXiv},
      primaryClass={cs.CL},
      url={https://arxiv.org/abs/1406.1078}, 
}

@misc{zerveas2020transformerbasedframeworkmultivariatetime,
      title={A Transformer-based Framework for Multivariate Time Series Representation Learning}, 
      author={George Zerveas and Srideepika Jayaraman and Dhaval Patel and Anuradha Bhamidipaty and Carsten Eickhoff},
      year={2020},
      eprint={2010.02803},
      archivePrefix={arXiv},
      primaryClass={cs.LG},
      url={https://arxiv.org/abs/2010.02803}, 
}

@article{Wu_2023,
   title={A comprehensive study of non-adaptive and residual-based adaptive sampling for physics-informed neural networks},
   volume={403},
   ISSN={0045-7825},
   url={http://dx.doi.org/10.1016/j.cma.2022.115671},
   DOI={10.1016/j.cma.2022.115671},
   journal={Computer Methods in Applied Mechanics and Engineering},
   publisher={Elsevier BV},
   author={Wu, Chenxi and Zhu, Min and Tan, Qinyang and Kartha, Yadhu and Lu, Lu},
   year={2023},
   month=jan, pages={115671} }

@article{https://doi.org/10.1002/cite.202100083,
author = {Schweidtmann, Artur M. and Esche, Erik and Fischer, Asja and Kloft, Marius and Repke, Jens-Uwe and Sager, Sebastian and Mitsos, Alexander},
title = {Machine Learning in Chemical Engineering: A Perspective},
journal = {Chemie Ingenieur Technik},
volume = {93},
number = {12},
pages = {2029-2039},
doi = {https://doi.org/10.1002/cite.202100083},
url = {https://onlinelibrary.wiley.com/doi/abs/10.1002/cite.202100083},
eprint = {https://onlinelibrary.wiley.com/doi/pdf/10.1002/cite.202100083},
year = {2021}
}

@misc{ramachandran2017searchingactivationfunctions,
      title={Searching for Activation Functions}, 
      author={Prajit Ramachandran and Barret Zoph and Quoc V. Le},
      year={2017},
      eprint={1710.05941},
      archivePrefix={arXiv},
      primaryClass={cs.NE},
      url={https://arxiv.org/abs/1710.05941}, 
}

@misc{kingma2017adammethodstochasticoptimization,
      title={Adam: A Method for Stochastic Optimization}, 
      author={Diederik P. Kingma and Jimmy Ba},
      year={2017},
      eprint={1412.6980},
      archivePrefix={arXiv},
      primaryClass={cs.LG},
      url={https://arxiv.org/abs/1412.6980}, 
}

@misc{wang2020understandingmitigatinggradientpathologies,
      title={Understanding and mitigating gradient pathologies in physics-informed neural networks}, 
      author={Sifan Wang and Yujun Teng and Paris Perdikaris},
      year={2020},
      eprint={2001.04536},
      archivePrefix={arXiv},
      primaryClass={cs.LG},
      url={https://arxiv.org/abs/2001.04536}, 
}

\end{document}